\documentclass[conference]{IEEEtran}
\IEEEoverridecommandlockouts
\usepackage{cite}
\usepackage{amsmath,amssymb,amsfonts}
\usepackage{graphicx}
\usepackage{textcomp}
\usepackage{xcolor}
\usepackage{algorithm}
\usepackage[noend]{algpseudocode}
\def\BibTeX{{\rm B\kern-.05em{\sc i\kern-.025em b}\kern-.08em
    T\kern-.1667em\lower.7ex\hbox{E}\kern-.125emX}}
\begin{document}

\title{Attentive Federated Learning for Concept Drift in Distributed 5G Edge Networks}

\author{\IEEEauthorblockN{Amir H. Estiri}
\IEEEauthorblockA{\textit{School of Computer Science} \\
\textit{McGill University}\\
Montreal, Canada \\
amir.estiri@mail.mcgill.ca}
\and
\IEEEauthorblockN{Muthucumaru Maheswaran}
\IEEEauthorblockA{\textit{School of Computer Science} \\
\textit{McGill University}\\
Montreal, Canada \\
maheswar@cs.mcgill.ca}
}

\maketitle

\begin{abstract}
Machine learning (ML) is expected to play a major role in 5G edge computing. Various studies have 
demonstrated that ML is highly suitable for optimizing edge computing systems as rapid mobility and 
application-induced changes occur at the edge. For ML to provide the best solutions, it is important to 
continually train the ML models to include the changing scenarios. 
The sudden changes in data distributions caused by changing scenarios (e.g., 5G base station failures) 
is referred to as concept drift and is a major challenge to continual learning. The ML models can 
present high error rates while the drifts take place and the errors decrease only after the model 
learns the distributions. This problem is more pronounced in a distributed setting where multiple ML models are being used for different heterogeneous datasets and the final model needs to capture all concept drifts. 
In this paper, we show that using Attention in Federated Learning (FL) is an efficient way of handling concept drifts. We use a 5G network traffic dataset to simulate concept drift and test various scenarios. The results indicate that Attention can significantly improve the concept drift handling capability of FL. 
\end{abstract}


\section{Introduction}
Edge computing (EC) is a new distributed computing idea that is inspired by cloud computing.
In EC, the compute and storage resources are pushed towards the edge (closer to end users)
of the network so that they are accessible with very small latency from end users. The main motivation for EC comes from data-intensive machine learning (ML) applications that are driven by Internet 
of Things (IoT) and want real-time responses from the ML hosted at the EC. While data-intensive ML
is a core application for EC, ML is also used by EC to optimize its own operations. In particular, the 
dynamicity created by vehicular networks and the 
time-varying demands created by the applications require EC to use sophisticated resource management schemes. Using ML to aid in resource management is one way of addressing the complexities presented by EC. \par
To handle the time varying nature of the EC problems, the ML models have to be continually updated
to keep them relevant for the changing
scenarios, which can alter the 
statistical properties of the data distributions 
in unpredictable ways. For example, when there is a festival in a city, the load at the edge server (sometimes referred to as fog servers or simply fogs) can be unexpectedly high and the requests 
can have a very different distribution from what is normally present. 
If the ML model is trained solely on request datasets from normal times, the ML model needs to be adapted to the new distribution posed by the festival-time requests -- this is an example of a temporary concept drift (we expect the request distribution to return to normal after the festival is over).  There are situations where concept drifts are permanent. For instance, the construction of a new building that blocks signals to a specific area reduces edge station coverage permanently causing a blind spot.
Concept drift is recognized as a major problem for continual learning~\cite{conc2}, especially in
large networks~\cite{conccont}. 
\par
Simple federated learning methods take a considerably long time to adapt the model in response 
to concept drift and reach a stable state. This happens because the weight averaging algorithm uses fixed coefficients regardless of a model experiencing drift or not. 
We propose using Federated Attention (FedAtt) \cite{fedatt}, a variant of the simple Federated Averaging (FedAvg) algorithm that uses dynamically altered coefficients to adapt to drift as 
quickly as possible.
\par
Our paper makes the following contributions with regard to concept drift in 
federated learning:
\begin{itemize}
    \item We show Attention in federated learning is a good way to handle concept drift.
    \item We show that the proposed method greatly reduces the time taken by a model to reach a stable state after concept drift.
    \item We test our model for single fog and multi fog concept drift scenarios caused by fog station failures.
    \item We devise a switching algorithm to handle situations where new fogs are introduced in edge networks.
    \item We show that the proposed method can adapt to temporary concept drifts without having to train a model from the beginning.
\end{itemize}
\par
This paper is structured as follows. In Section 2,
we present an overview of the existing approaches proposed for handling concept drift in EC
systems. We divide Section 3 into multiple scenarios that motivate the use of our proposed model. In Section 4, we formulate the problem thoroughly by explaining the proposed attentive federated algorithm. In Section 5, we explain the dataset, model architectures and model setups. Then, we conduct multiple experiments to test our system in different scenarios and settings. We show the results of our experiments for the proposed model along with the baseline model for comparison and prove that our method is superior in dealing with concept drift. In the end, we conclude the work and look more deeply into the areas that can still be explored to further improve this work.

\section{Related Works}
Traditionally, ML models are trained on a single system or clusters by centralizing data from distributed sources. In many applications, this requires a prohibitive amount of data communication. As a solution, it was suggested to locally train models and only average the model weights. McMahan et al. \cite{f25} empirically evaluated deep learning model averaging in a decentralized layout and coined the term Federated Learning.
\par
However, little research has been done regarding the scenario where data distribution is non-identical among the participants and also changes over time in unforeseen ways, causing what is known as concept drift. This situation is very common in real-life applications and poses new challenges to both federated and continual learning \cite{conccont}.
\par
Many methods have been proposed to battle the problem of concept drift in federated settings. Solutions have been proposed to mitigate the need for constant periodic updates and invest communications only when necessary \cite{dynavg}. Dynamic Averaging is one example of an alternate to federated averaging that optimally communicates model weights only when model divergence exceeds a threshold. However, dynamic model averaging only works best when drifts are scarce in time. This solution intends to optimally focus communications over network when drift occurs. But in a large distributed fog network which is in constant communication with multiple edge devices, concept drift can be a common phenomenon and a continual approach is needed. Even though Dynamic Averaging reduces communications on the network, the model is still helpless when concept drift starts happening. There is still a huge delay between the moment drift happens, and the moment model adapts to the new data and reaches a stable state.
\par
Other methods such as Gossip-based learning framework \cite{d} detect and handle concept drift in peer-to-peer (P2P) networks. This algorithm does not collect data at a central location, instead it uses a handful of online learners taking random walks in the network \cite{d6} and training on each node's dataset at each step. Lack of a global model in P2P based methods prevents us from employing this architecture. Vehicles traveling in an area need fast access to the network map from every edge server. Therefore, every edge server should have fast access to a global model and that model should perform with great accuracy on every edge server data. Similar works \cite{d2} \cite{d3} have proposed methods to handle drifts in distributed P2P networks, but still lack a central model that captures the distribution of the entire data over different edge servers.
\par
The most important part of federated learning is the federated optimization on the server side which aggregates the client models. In this paper, we use a self-adaptive federated optimization strategy to aggregate ML models from decentralized clients. We call this Attentive Federated Aggregation, Federated Attention or FedAtt for short. It was first introduced by \cite{fedatt} for learning private neural language models on distributed mobile device keyboards. It introduces the attention mechanism for federated aggregation. This model improves upon the novelty by reducing the time between drift happening and system reaching a stable state. To the extent of our knowledge, no previous work has focused on this aspect of the concept drift problem in federated learning.

\section{Motivating Scenarios}\label{sec:motiv}
In this section, we present several concept drift scenarios. We construct experiments to evaluate concept drift mitigation under these scenarios.

\subsection{Scenario 1 - Faulty Station}
Probably the most frequent cause of concept drift in distributed edge applications is when a base station goes offline. This could have many causes, from high network load to faulty hardware. Adapting to this concept drift is probably the most important feature of a system such as ours. When a base station goes offline, other stations close to the offline station pick up the extra load. The extra load causes a shift in input distribution for the nearby edge servers.

\subsection{Scenario 2 - Catastrophic Forgetting}
In order to achieve adaptive models, federated learning deals with two conflicting objectives: retaining previously learned knowledge that is still relevant, and replacing any obsolete knowledge with current information. This is usually known as the stability-plasticity dilemma \cite{new1-9}. Continual learning poses particular challenges for artificial neural networks due to the tendency for knowledge of previously learnt task to be abruptly lost as information relevant to the current task is incorporated. This phenomenon, termed catastrophic forgetting \cite{new1-11}, occurs when we are dealing with concept drift. The model has to quickly adapt itself to new knowledge and keep previous knowledge for as long as it is needed.

\subsection{Scenario 3 - Knowledge Transfer}
An efficient concept drift management strategy should save important information, as well as being able to transfer knowledge and skills to future tasks. This falls within the transfer learning paradigm \cite{new2-13}, which focuses on storing knowledge gained while solving one problem and applying it to a different but related problem. In this case, a global ML model obtained in the cloud would be a general solution that could be then adapted to each device locally. One example of this situation is when a specific pattern of concept drift happens to different areas of the city. The system should be able to transfer its knowledge across areas and adapt to the same concept drift, as it has been observed in another fog previously.

\subsection{Scenario 4 - New Station}
One scenario that has the lowest chance of happening but a long-lasting impact is when a new base station is introduced in the area. This event can happen more frequently in 5G compared to other networks, due to the denser deployment of stations. When a new station appears, a new model has to be trained on the data stream. When the model has been trained on enough data and is ready to be shared with the global model. After several rounds of communication, the system achieves a stable state. It takes a long time to train a model on all previous data. Therefore, during training, there is no model to use for inference. Due to tight constraints on latency in an edge computing architecture, we can not collect data and train a model from scratch. We propose using a classifier to cluster the incoming queries and use the previous local models as backup.

\section{Proposed Method}
Now that we have made the significance of concept drift clear in ML and especially federated learning, we will explain the proposed method. We propose changing the aggregation algorithm in federated averaging such that the aggregation converges faster to the optimum error rate. We use attention to change the aggregation weights dynamically and achieve this objective.
\par
The motivation behind using attention is that, it gives system the ability to actively change the contribution of each local model, based on how much it contributes to bringing us closer to the main objective. As opposed to the naive Federated Averaging (FedAvg) algorithm that averages models with fixed aggregation weights. The main objective here is to have a global model that is close to all local models as possible in weight space and therefore has the best accuracy possible on all local datasets. The optimization loss function for the aggregation can be written as:

\begin{equation}\label{eq:loss}
    \mathcal{L} = \sum_{k=1}^m \frac{1}{2}\alpha_k L(w_t,w^k_{t+1})
\end{equation}

where $w_t$ is the parameters of the global server model at time $t$, $w^k_{t+1}$ is the parameters of $k$-th local model at time $t+1$. $L(. , .)$ is defined as the distance between two sets of neural parameters, and $\alpha_k$ is the attention weight vector to measure the importance of weights for the client models.
\par
Unlike the popular attention mechanism applied to the data flow (model input), the attentive aggregation in our method is applied on the learned parameters (model weights) of the neural network. Given the parameters in the $l$-th layer of the server model denoted as $w^l$ and parameters in the $l$-th layer of the $k$-th client model denoted as $w^l_k$ , the similarity between global and $k$-th local model in the $l$-th layer is calculated as the $L2$ norm of the difference between two vectors and passed through a softmax layer to acquire the attention vectors.

\begin{equation*}
    s^l_k = ||w^l-w^l_k||_2
\end{equation*}

\begin{equation*}
    \alpha^l_k = softmax(s^l_k) = \frac{e^{s^l_k}}{\sum_{k=1}^m e^{s^l_k}}
\end{equation*}

The federated attention mechanism on the parameters works in a layer-wise fashion. There are attention scores for each layer in the neural networks. For each model, the attention score is $\alpha_k = \{\alpha^0_k, \alpha^1_k,...,\alpha^l_k,...\}$. Using the L2-distance for $L(. , .)$ and taking the derivative of the objective function in Equation \ref{eq:loss}, we get the gradient and can perform gradient descent on the loss objective.

\begin{equation*}
    \nabla_{w_t} \mathcal{L} = \sum_{k=1}^m \alpha_k (w_t - w^k_{t+1})
\end{equation*}

\begin{equation*}
    w_{t+1} \gets w_t - \epsilon \nabla_{w_t} \mathcal{L}
\end{equation*}

where $\epsilon$ is the step size. The full procedure of this optimization algorithm is described in Algorithm \ref{alg:fedatt}. It takes the server parameters $w_t$ at time $t$ and client parameters $w^1_{t+1},...,w^m_{t+1}$ at time $t+1$, and returns the updated parameters of the global server.

\begin{algorithm}
\caption{Attentive Federated Averaging}\label{alg:fedatt}
\begin{algorithmic}[1]
\State $k$ is the ordinal of clients;$l$ is the ordinal of neural layers;$\epsilon$ is the step-size of global model optimization
\State \textbf{Input}: server parameters $w_t$ at time t,client parameters $w^1_{t+1},...,w^K_{t+1}$ at $t+1$
\State \textbf{Output}: aggregated server parameters $w_{t+1}$ at $t+1$
\Procedure{ATTENTIVE OPTIMIZATION}{$w_t$,$w^k_{t+1}$}
\State Initialize $\alpha = \{\alpha_1,...,\alpha_m\}$\Comment{Attention for each clients}
\For{each layer $l$ = $1$, $2$, $...$}
\For{each user $k$}
\State $s^l_k = ||w^l-w^l_k||_2$
\EndFor
\State $\alpha^l_k = softmax(s^l_k) = \frac{e^{s^l_k}}{\sum_{k=1}^K e^{s^l_k}}$
\EndFor
\State $w_{t+1} \gets w_t - \epsilon \sum_{k=1}^K \alpha_k (w_t - w^k_{t+1})$
\State \textbf{return} $w_{t+1}$
\EndProcedure
\end{algorithmic}
\end{algorithm}

\begin{figure*}
    \centering
    \includegraphics[width=0.9\textwidth]{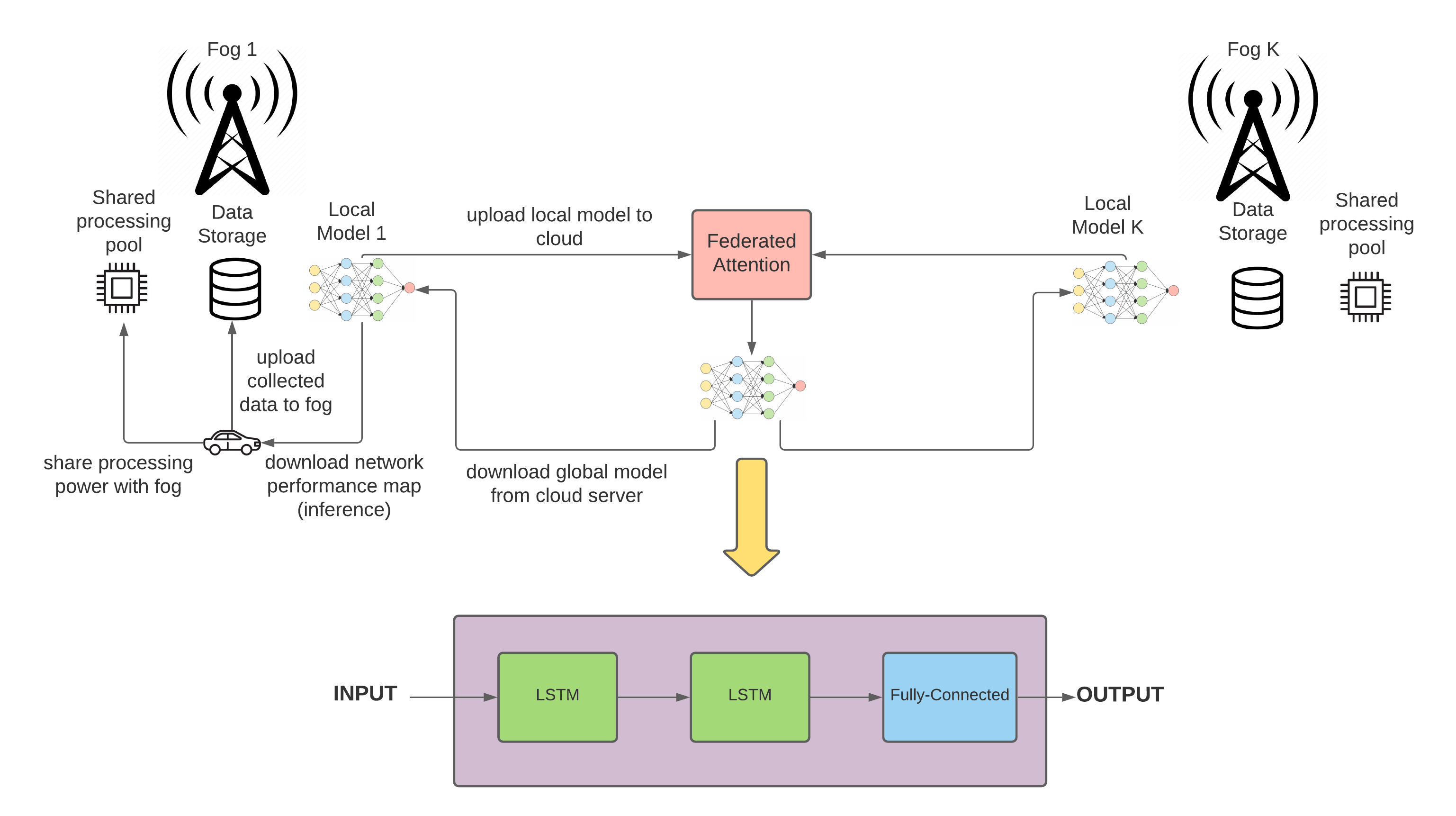}
    \caption{System diagram for the federated attention}
    \label{fig:sys-diag}
\end{figure*}

\section{Experiments \& Analysis}
In this section we conduct multiple experiments using the 5G dataset that are designed to test the adaptability of the FedAtt model and compare it to the baseline FedAvg. We explain the model architecture, dataset and analyze different scenarios of concept drift as explained in Section \ref{sec:motiv}

\subsection{Dataset}
The dataset \cite{5gdataset} consists of 5G trace data collected from a major Irish mobile operator in 2020. The dataset has two mobility patterns (static and dynamic), and across two application patterns (video streaming and file download). The dataset is composed of client-side cellular key performance indicators (KPIs) comprised of channel-related metrics, context-related metrics, cell-related metrics, throughput information and latency values.
\par
To create drift in our dataset, we synthetically alter throughput values for some fogs after a specific point of time. To simulate, we simply add a constant value ($0.5$ in this case) to the throughput values. To show our model's performance, we monitor the Mean Absolute Error (MAE) for all client models during the experiment.

\subsection{Model Architecture}
Many ML applications in edge networks are concerned with predicting temporal characteristics of network data (bandwidth, latency, throughput, etc). Therefore, for our local model architecture, we decided to use a recurrent neural network model due to their ability in processing temporal data. The model consists of two Long-Short Term Memory (LSTM) layers and a Fully-Connected layer to predict throughput for a 5G network. We use the 5G dataset, distribute it across all fogs and use the location and time values to build a prediction function for the throughput values. We use this as the architecture for our local and global models and conduct the experiments on the drifted 5G dataset.

\subsection{Significance of Concept Drift in Federated Systems}
\begin{figure}
    \centering
    \includegraphics[width=0.6\linewidth]{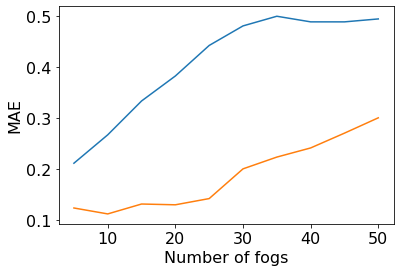}
    \caption{Variation of MAE with different number of fogs for FedAtt (Orange) and FedAvg (Blue)}
  \label{fig:numfogs}
\end{figure}

To understand the extent to which FedAvg and FedAtt deal with concept drift in federated learning, we conducted multiple experiments with different number of fogs. The average MAE across all fogs is depicted in Figure \ref{fig:numfogs} for both FedAtt and FedAvg.
\par
This experiment was done for global epochs on the 5G dataset for different number of fogs ($K$). Results show that the more fogs we have, the longer it takes for the federated averaging algorithm to converge to lower errors compared to federated attention.

\subsection{Single-Fog Drift Adaptability}
In the next experiment, we introduce concept drift to only one fog station. The purpose of this experiment is to observe how the attentive federated learning method is able to notice and adapt to concept drifts happening on a single node in the distributed network.
\par
\begin{figure}
  \centering
  \begin{minipage}[b]{0.49\linewidth}
    \includegraphics[width=\linewidth]{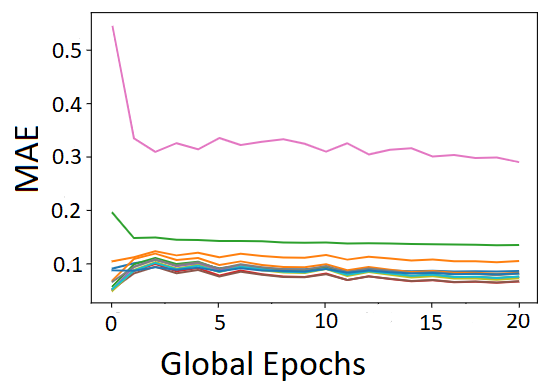}
  \end{minipage}
  \hfill
  \begin{minipage}[b]{0.49\linewidth}
    \includegraphics[width=\linewidth]{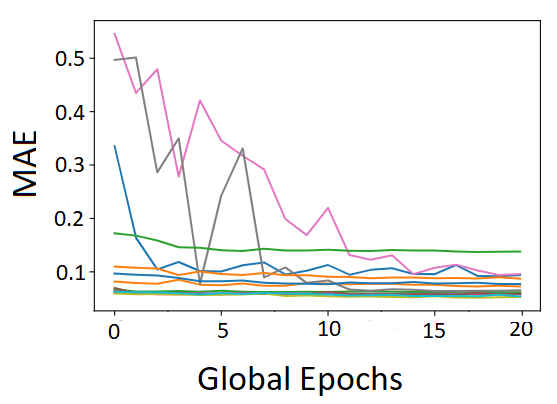}
  \end{minipage}
  \caption{MAE error for Federated Averaging (left) and Federated Attention (right) with drift - Pink line shows the error of the (single) drifted fog}
  \label{fig:fedexp1}
\end{figure}

The experiment was conducted on a simple federated learning architecture without attention (FedAvg). The result of the error is shown with respect to global epoch numbers. As can be seen in the results of Figure \ref{fig:fedexp1}, the drifted fog station (pink line) can not adapt itself to concept drift even after $20$ epochs and the error reaches a constant value of $0.35$ which is very high. At the same time all other fog stations have achieved low errors of $<0.15$.
\par
The same experiment is conducted on the same dataset, with the same models and settings. The only difference is that we added a self-adaptive attention mechanism in the aggregation phase. As can be seen from Figure \ref{fig:fedexp1}, the same fog station that was unable to adapt to concept drift, has now achieved MAE error of $0.1$ in $20$ global epochs. This result proves the effectiveness of FedAtt in dealing with single fog concept drifts in distributed edge networks.

\subsection{Temporary Drift Adaptability}
Now, we want to check if FedAtt can react to concept drifts in single fog stations and still remember the pre-drift data distribution without having to retrain the model.
\par
In this experiment we introduce the same concept drift as before. But we revert the data distribution back to the normal situation after some time. The purpose of this experiment is to show that FedAtt can react to new concept drifts without completely discarding its previous information of the system (Catastrophic Forgetting).
\par
\begin{figure}
  \centering
  \begin{minipage}[b]{0.49\linewidth}
    \includegraphics[width=\linewidth]{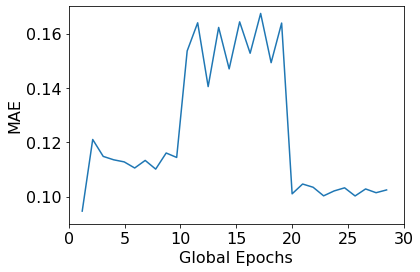}
  \end{minipage}
  \hfill
  \begin{minipage}[b]{0.49\linewidth}
    \includegraphics[width=\linewidth]{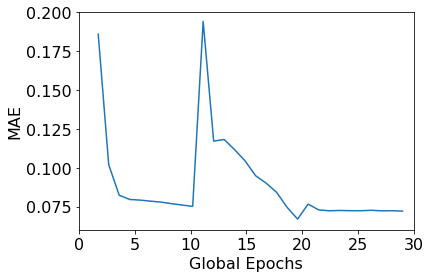}
  \end{minipage}
  \caption{Temporary drift - Federated Averaging model (left) does not discard previous knowledge of the distribution after learning on drifted data, however it can not adapt to the temporary drift fast enough. Federated Attention (right) model does not discard previous knowledge of the distribution after learning on drifted data}
  \label{fig:exp1}
\end{figure}

The experiment is divided into three time stages. In the first and third stages we use the normal data distribution, but in the second stage we introduce drift to one of the fogs. The result of the experiment for FedAtt and FedAvg are shown in Fig \ref{fig:exp1}.
\par
FedAtt can adapt to drift much better than FedAvg, even though both models can remember temporary drifts. This proves that FedAtt can pick up temporary drift patterns without the need to be retrained again after reverting back to the original distribution and also adapts to the drift much faster than FedAvg. Scenarios with temporary drifts in fog computing can happen quite frequently. For example, temporary station failures or unexpected increase in network traffic are just a few of the many events that can lead to this situation.

\subsection{Drift Adaptability Over Different Fogs}
The objective of this experiment is to check how well FedAtt adapts to a specific type of drift over different fog nodes (Knowledge Transfer). In this experiment, we add drift to one fog station for a period of time. Then we add the same concept drift to another fog station at a different time and observe how the system can translate its adaptability to drift, across fogs in the network.
\par
\begin{figure}
  \centering
  \begin{minipage}[b]{0.49\linewidth}
    \includegraphics[width=\linewidth]{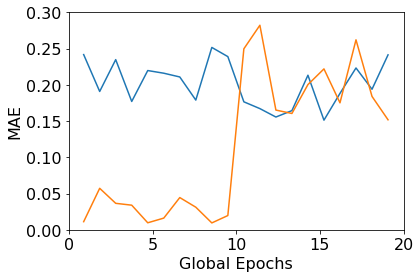}
  \end{minipage}
  \hfill
  \begin{minipage}[b]{0.49\linewidth}
    \includegraphics[width=\linewidth]{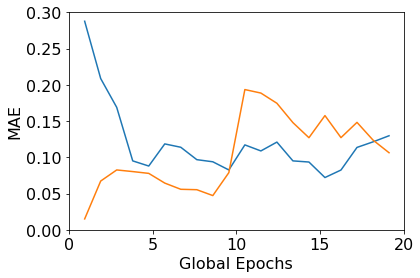}
  \end{minipage}
  \caption{Federated averaging (left) can not adapt to previously encountered drifts happening across different fog nodes. Attentive federated (right) adapts to drifts happening on different fog nodes after having encountered the same drift distribution in another fog at a previous time}
  \label{fig:exp2}
\end{figure}

To precisely design this experiment, we choose two fogs which we call blue and orange. We drift blue from epoch $0$ and drift orange from epoch $10$. The error for both fogs are shown at each step in Fig \ref{fig:exp2}.
\par
The FedAtt plot shows that the blue model adapts to drift after $10$ steps. When the orange model encounters the same concept drift, it experiences a low increase in error and adapts to the concept drift very fast. On the other hand, for FedAvg, when orange encounters the same drift, it has a much larger error. This shows FedAtt's ability to learn drift patterns across different fogs.

\subsection{Multi-Fog Drift Adaptability}
\begin{figure}
  \centering
  \begin{minipage}[b]{0.49\linewidth}
    \includegraphics[width=\linewidth]{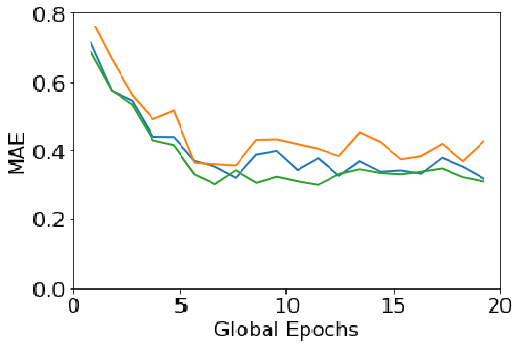}
  \end{minipage}
  \hfill
  \begin{minipage}[b]{0.49\linewidth}
    \includegraphics[width=\linewidth]{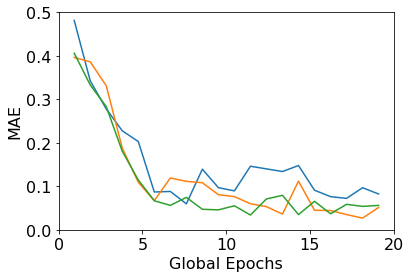}
  \end{minipage}
  \caption{Federated averaging (left) can not adapt to drifts in multiple different fog nodes. Attentive federated model (right) reacts to drifts in $3$ different fog nodes}
  \label{fig:exp3}
\end{figure}

For this experiment, we want to show how FedAtt can adapt itself to several concept drifts at different fog stations at the same time. We design an experiment similar to the single-fog drift experiment. But we drift two other fog stations along with the first one. Fig \ref{fig:exp3} shows the results of both algorithms under this experiment. The results show that FedAtt can even adapt to multiple (in this case $3$) concept drifts in distributed networks.

\subsection{New Fog Stations}
The moment a new base station is added to the network, all nearby fogs will be affected. The new fog starts collecting data and training its local model. However, in the meantime the new fog can not accommodate user queries with an accurate answer. We design a switching algorithm that uses a two-layer Fully-Connected layer to classify the incoming queries and feed them into the correct pre-drift fog. The system will switch to its normal procedure after the new fog reaches a stable state. We design a multi-class classification neural network. We train the model with cross-entropy loss function and check the model performance on a test dataset. 
\par
\begin{figure}
  \centering
  \begin{minipage}[b]{0.49\linewidth}
    \includegraphics[width=\linewidth]{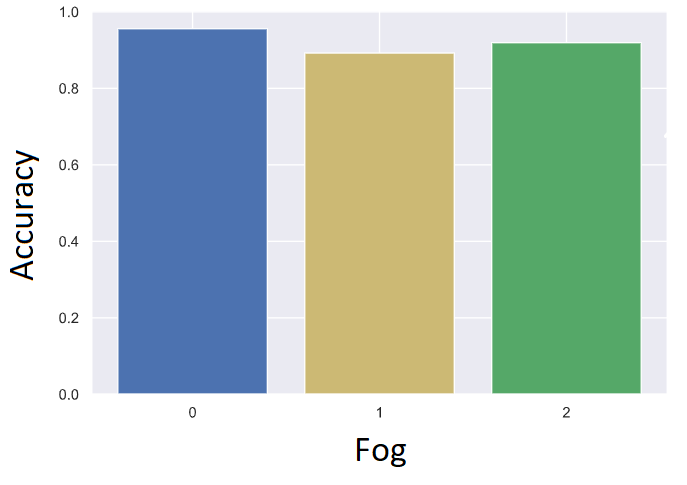}
    \caption{Accuracy for classification in time-series classification} 
    \label{fig:dtw}
  \end{minipage}
  \hfill
  \begin{minipage}[b]{0.49\linewidth}
    \includegraphics[width=\linewidth]{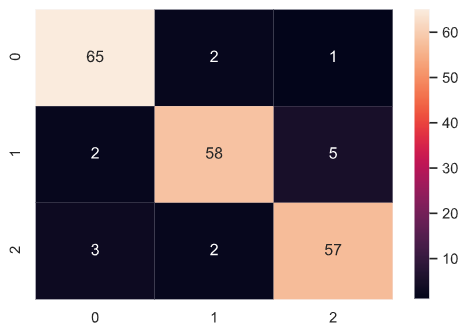}
    \caption{Confusion Matrix for time-series classification}
    \label{fig:confmat}
  \end{minipage}
\end{figure}

We introduce a new fog in the 5G dataset. The coverage area for this new fog is surrounded by $3$ fogs, each of which have their own local model trained. We test our time-series classification model by dividing the dataset into a train and test set for all $3$ nearby fogs. We then classify the incoming data in the test set. Fig \ref{fig:dtw} shows the classification accuracy for each fog station, along with the confusion matrix in Fig \ref{fig:confmat}. The results show that a small classification neural network can be used to temporarily deal with user requests when a new station appears.

\section{Conclusion \& Future Works}
We proposed using attention for federated aggregation to make distributed continual learning models more adaptive to concept drifts. We used a real-world 5G trace dataset and proved that our model responds very well to concept drifts, which is a common phenomena in distributed networks and continual learning models. Our proposed model achieved nearly $20\%$ less mean absolute errors than the baseline federated averaging algorithm in single and multiple fog drift scenarios. We also constructed a switching algorithm to classify queries from nearby fogs when a new fog is introduced in an area and achieved $95\%$ accuracy. We also showed that federated attention can handle temporary drifts and transfer knowledge over different local fog models.
\par
Further research works could focus on methods that ensure low errors for all fog nodes after each communication round. Our proposed method ensures that after a number of global epochs, error values for all fog stations reach small values. But it makes no such guarantees for the accuracy while in the process of adapting to concept drifts. This opens up new avenues for further research and investigation.


\vspace{12pt}

\end{document}